\documentclass[runningheads]{llncs}

\usepackage{eccv}

\usepackage{eccvabbrv}

\usepackage{graphicx}
\usepackage{booktabs}
\usepackage{color}
\usepackage{colortbl}
\usepackage{multirow}
\usepackage{array}
\newcommand{\myparagraph}[1]{\noindent\textbf{#1}}

\usepackage[accsupp]{axessibility}  

\usepackage[pagebackref,breaklinks,colorlinks,citecolor=eccvblue]{hyperref}

\usepackage{orcidlink}

\begin{document}

\title{Video Editing for Video Retrieval} 

\titlerunning{Video Editing for Video Retrieval}

\author{Bin Zhu \inst{1}$^*$\orcidlink{0000-0002-9213-2611} \and
Kevin Flanagan \inst{2} \and 
Adriano Fragomeni \inst{2} \and\\ 
Michael Wray \inst{2}\orcidlink{0000-0001-5918-9029} \and
Dima Damen \inst{2}\orcidlink{0000-0001-8804-6238}
}

\renewcommand{\thefootnote}{\fnsymbol{footnote}}
\footnotetext[1]{Work carried out while Bin Zhu was employed at the University of Bristol.}

\authorrunning{B. Zhu et al.}

\institute{Singapore Management University \and
University of Bristol}

\maketitle

\begin{abstract}
     Though pre-training vision-language models have demonstrated significant benefits in boosting video-text retrieval performance from large-scale web videos, fine-tuning still plays a critical role with manually annotated clips with start and end times, which requires considerable human effort. To address this issue, we explore an alternative cheaper source of annotations, single timestamps, for video-text retrieval. We initialise clips from timestamps in a heuristic way to warm up a retrieval model. Then a video clip editing method is proposed to refine the initial rough boundaries to improve retrieval performance. A student-teacher network is introduced for video clip editing: the teacher model is employed to edit the clips in the training set whereas the student model trains on the edited clips. The teacher weights are updated from the student's after the student's performance increases.
    
    Our method is model agnostic and applicable to any retrieval models. We conduct experiments based on three state-of-the-art retrieval models, COOT, VideoCLIP and CLIP4Clip. Experiments conducted on three video retrieval datasets, YouCook2, DiDeMo and ActivityNet-Captions show that our edited clips consistently improve retrieval performance over initial clips across all the three retrieval models.
\end{abstract}

 \section{Introduction}
\label{sec:intro}

With the proliferation of online videos, the significance of video-text retrieval has grown in the context of video search and navigation. 
Particularly for assistive systems, the ability to retrieve a relevant part of a long video, or a clip from a large corpus is crucial for sifting through the exponential growth of videos available online. 
Text-to-video retrieval has demonstrated notable advancements in performance in recent years, primarily attributed to the emergence of vision-language pre-training methodologies~\cite{miech2020end, xu2021videoclip, lei2021less, xue2023clipvip, lavila} and cross-modal fusion techniques~\cite{liu2019use, coot, clip4clip, wray2019fine, XPool, wu2023cap4video}. The efficacy of these retrieval models heavily depends on the availability of meticulously annotated retrieval datasets, where precise temporal boundaries are defined for each caption, such as MSR-VTT~\cite{Msr-vtt}, YouCook2~\cite{ZhXuCoAAAI18}, ActivityNet-Captions~\cite{krishna2017dense}.

\begin{figure}
    \centering
    \includegraphics[width=0.75\linewidth]{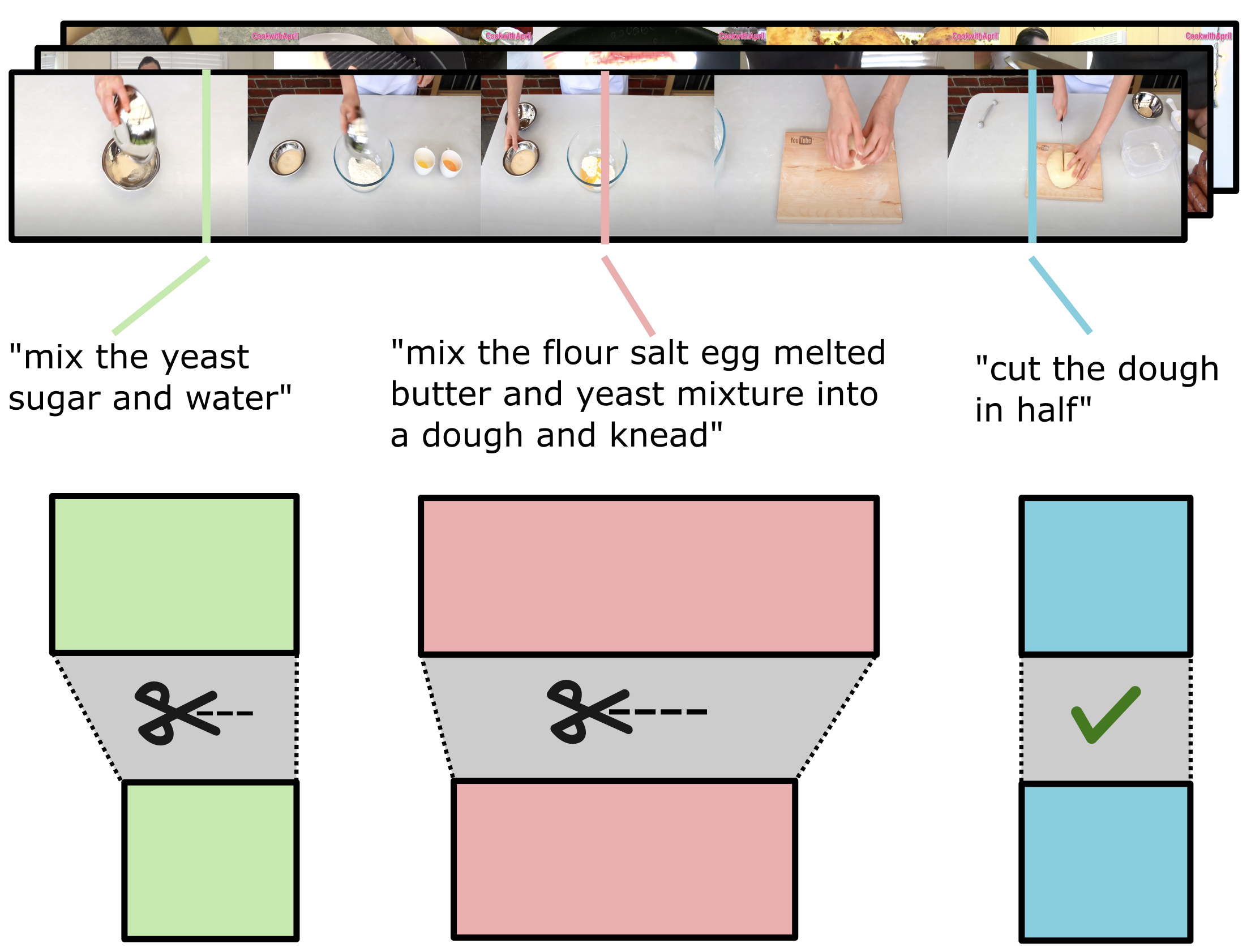}
    \caption{Given videos with rough timestamp supervision, we propose a method to edit clips increasing video retrieval performance.}
    \label{fig:intro:fig}
\end{figure}

The advantages of pre-training on large-scale, webly-supervised video datasets \cite{miech2020end,xu2021videoclip,shvetsova2022everything} have been shown to significantly enhance retrieval performance. Nevertheless, it is essential to acknowledge that such datasets tend to exhibit a degree of noise due to the generation of captions through Automatic Speech Recognition (ASR) systems, with estimates indicating that approximately 30\% to 50\% of the data could be noisy~\cite{miech2020end,han2022temporal}. Consequently, while these large-scale datasets are instrumental for pre-training purposes, the inclusion of manually annotated video clips remains crucial for the fine-tuning phase, ensuring the attainment of compelling retrieval performance.

In this paper, we investigate an alternative approach to weakly supervised annotations for text-to-video retrieval. Instead of relying on fully supervised data that provides precise start and end times for each caption, we explore a more limited form of annotation, which consists of a single timestamp associated with each caption. In this scenario, captions do correspond to parts of the video, roughly around the labelled timestamp. This has the considerable benefit of reducing annotation time---~\cite{ma2020sf} found annotating single timestamps to be 6$\times$ faster than annotating start/end times. In the literature, single timestamp annotations have been already explored in the tasks of temporal action localisation~\cite{ma2020sf, lee2021learning} and video moment retrieval~\cite{cui2022video, flanagan2023learning}. Nevertheless, to our best knowledge, how to improve the video-text retrieval performance using single timestamps is not explored yet.

We address this issue by introducing a video editing paradigm for video-text retrieval. Leveraging the single timestamps, rough start and end times of a clip for each caption can be initialised by heuristics. The primary objective of this paper is to refine the rough temporal boundary by video clip editing, ultimately leading to improved text-to-video retrieval performance. Our proposed method is composed of two key stages: warm-up and co-training.
Specifically, in the first stage, we construct an initial set of clip-caption pairs, leveraging heuristics based on single timestamps to train a retrieval model. These initial clips with rough start and end times corresponding to the caption can still offer a valuable supervisory signal, facilitating the learning of a shared embedding space between captions and video clips. 

The second stage introduces a student-teacher network, co-trained concurrently to perform video clip editing and video retrieval. Both the teacher and student networks share a common architecture and are initialized with the weights derived from the warm-up retrieval model. The teacher network refines the initial clip boundaries by identifying the most semantically similar and representative clips related to a given caption. In contrast, the student network is trained using the edited clips generated by the teacher network, with the aim of enhancing retrieval performance. It's important to note that the parameters of the student network are updated in each training iteration, while the teacher network receives updates based on the performance improvements observed in the student network. This co-training process continues until convergence is achieved.

We conduct experiments with three retrieval models, including COOT~\cite{coot}, VideoCLIP~\cite{xu2021videoclip} and CLIP4Clip~\cite{clip4clip} on three commonly used video retrieval datasets: YouCook2~\cite{ZhXuCoAAAI18}, DiDeMo~\cite{DiDeMo} and ActivityNet-Captions~\cite{krishna2017dense}. By incorporating our proposed video clip editing method, we observe consistent improvements in retrieval performance across all three datasets using the three retrieval models. Furthermore, human study showcases that our edited clips align better with human perception than the initial clips. We summarise our contributions as follows:
(i) 
(i) We introduce the first work to improve the video-text retrieval performance only using single timestamps as annotations. We propose a model-agnostic approach that enables to refine the start and end times initialised from single timestamps.
(ii) We demonstrate that training retrieval models with our edited clips leads to enhanced clip retrieval performance across all three datasets, employing three distinct retrieval models.
(iii) We conduct a comprehensive analysis and ablation study of our proposed method on the YouCook2 dataset.

\section{Related work}
\label{sec:related}


\subsection{Text-to-video Retrieval}
Text-to-video retrieval aims to retrieve corresponding videos by giving a text query. One line of works focus on developing techniques to tackle cross-modal fusion or alignment~\cite{liu2019use, gabeur2020multi, coot, clip4clip, wray2019fine, fragomeni2022contra, XPool, hu2022lightweight, wu2023cap4video}. 
In contrast, another line of works have leveraged the power of pre-training using large-scale datasets to achieve impressive performance~\cite{miech2020end, xu2021videoclip, lei2021less, kevin2022egovlp, shvetsova2022everything, xue2023clipvip, lavila, cheng2023vindlu, egovlpv2}.

Our work differs from these works where each caption is annotated with start and end times. More related is the recent work EgoVLP~\cite{kevin2022egovlp, egovlpv2} where start and end times are obtained from single timestamps within untrimmed videos using a heuristic. In our work, we learn-to-edit the start and end times initialised from single timestamps to improve clip retrieval performance. 

Note that we address the task of text-to-video retrieval in this paper, albeit training using single-timestamp supervision. This differs from the video-moment retrieval task which finds a specific `moment' of a video given a query~\cite{anne2017localizing, gao2017tall, mithun2019weakly}. Recently, single-timestamp annotation is explored in~\cite{cui2022video} for video-moment retrieval using contrastive learning. In contrast, we focus on editing clips in untrimmed videos to improve clip retrieval performance. 

\subsection{Learning from Noisy data}

Our work is related to previous efforts to correct noisy or incorrect labels for improving performance.
Most works focus on dealing with noisy labels in classification tasks~\cite{yi2019probabilistic, xiao2015learning, song2022learning}. The existing efforts range from sample re-labelling~\cite{reed2014training, tanaka2018joint} to re-weighting samples~\cite{jiang2018mentornet, ren2018learning}. Although great progress has been achieved in learning with noisy labels, correcting cross-modal correspondences (e.g., vision and text) is rarely explored. In~\cite{cross-modal-noisy}, a Noisy Correspondence Rectifier (NCR) is proposed for image and text matching when learning from noisy pairs. To mitigate the impact of noise, NCR first divides the data into clean and noisy sets based on the sample loss for each pair, then dynamically assigns a softer margin for the triplet loss in order to achieve robust cross-modal retrieval. 

Temporal Alignment Network (TAN)~\cite{han2022temporal} is proposed to predict whether the caption and clip are alignable or not. If alignable, the temporal alignment of the clip is corrected. WYS$^2$~\cite{ashutosh2023you} further explores visual narration detection. WYS$^2$ aims to determine whether the actions in the video can be described by the narrations. As TAN only focuses on the relevance between the caption and clip, WYS$^2$ imposes a further constraint that the caption can visually describe the action in the video clip. As reported in TAN, the employed dataset HowTo100M~\cite{howto100m} contains only 30\% visually aligned clip-caption pairs. The problems of both TAN and WYS$^2$ are different from ours, where we work with untrimmed datasets with 100s of captions per video, however, the caption \emph{always} corresponds to a part of that video. In particular, we focus on the case that captions are narrated by human annotators to describe the actions in the video.
Another distinction between our work and that in~\cite{han2022temporal} is that we can change the length of the clip, while in~\cite{han2022temporal} the duration of the clip is \emph{assumed optimal}.
We showcase the benefit of changing the clip's duration 
for caption-to-clip retrieval.


\begin{figure*}[ht]
    \centering
    \includegraphics[width=\textwidth]{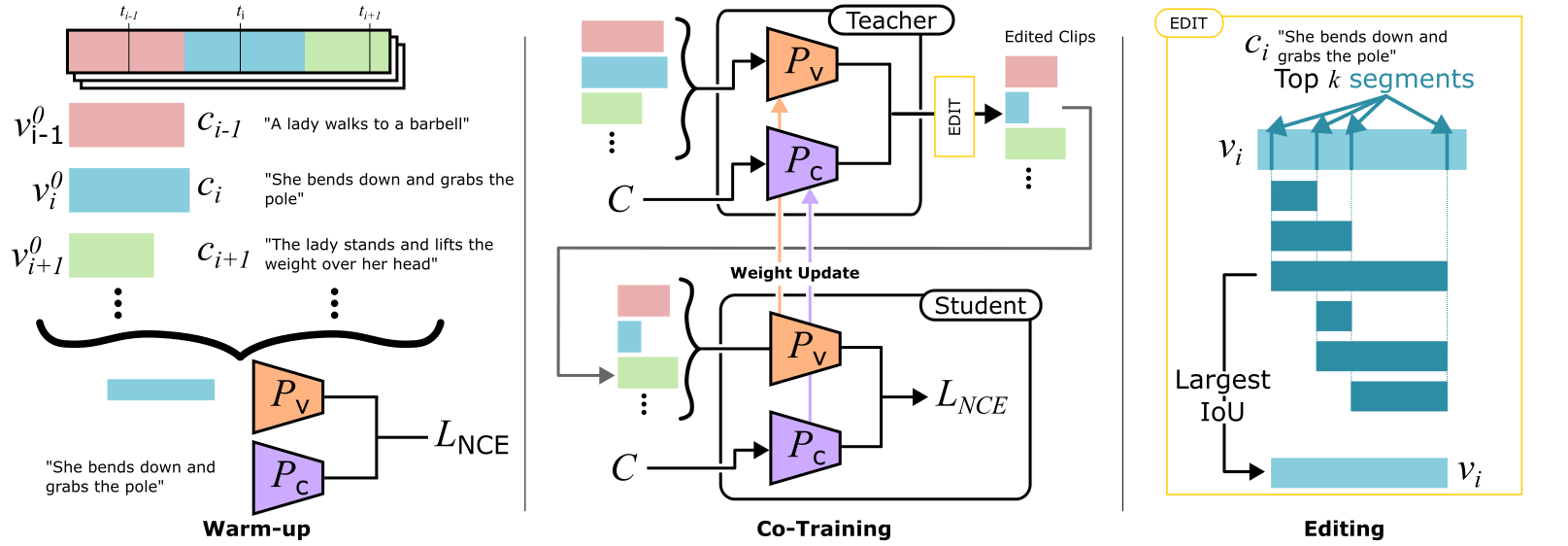}
    \caption{Overview of the method. Left, we warm up a text-video retrieval model from initial clips $v_i^0$ in the first stage, $P_v$ and $P_c$ refer to clip and caption encoders respectively. Centre, a teacher edits clips to have maximum similarity with their corresponding captions. A student model learns on the edited clips and once performance increases, the teacher's weights are updated with that of the student's. Right, we edit a video by selecting the top $k$ similar segments with the corresponding caption, candidates are creating using these as start/end times and the clip with the highest average IoU is chosen as the new clip.}
    \label{fig:method}
\end{figure*}

\section{Method}
\label{sec:method}


\subsection{Problem Definition}
\label{sec:def}
From a text query, text-to-video retrieval methods search for the corresponding video within a video gallery. To differentiate from paragraph-to-video retrieval~\cite{zhang2018cross}, we focus on caption-to-clip retrieval, where clips are trimmed from the videos leveraging start and end time annotations by human. In this paper, we consider the start and end times are not available to trim the videos into clips, instead, only a single timestamp for each caption is provided. 
The goal of this paper is to edit the video clips in order to refine the rough boundaries initialised from single timestamps, so as to improve the caption-to-clip retrieval performance.

Given a set of untrimmed videos $\mathcal{V}$, annotated by a total of $M$ captions across all videos. We consider the set of all captions, and their associated single timestamps such that, $\mathcal{C} = \{(c_i, t_i)\}_{i=1}^{M}$, where the $i$-th caption $c_i$ corresponds to timestamp $t_i$. 
In line with previous works such as~\cite{howto100m, kevin2022egovlp}, the videos are trimmed into rough clips first. Given three consecutive timestamps ${(t_{i-1}, t_i, t_{i+1})}$ in the video, we define the video clip for caption $c_i$ based on the midpoint between the neighboring timestamps, i.e.,~[${s_i=\frac{t_{i-1}+t_{i}}{2}}$, $e_i=\frac{t_{i}+t_{i+1}}{2}$]. We refer to this initial clip as $v_i^0$. We explore alternative options to form the initial clips in Sec~\ref{subsec:ablation}. The initial clips can also be derived from other sources of annotations, for example, potentially noisy human annotations. Our method then takes these initial clip-caption pairs and edits them to improve clip-retrieval performance.

\subsection{Warm-up and Co-Training}
\label{sec:cotraining}

\paragraph{Overview.}
The overall framework of our proposed method is illustrated in Figure~\ref{fig:method}. Our method is composed of two stages: warm-up and co-training. In the first stage, the initial clip-caption paired data is used to train a cross-modal clip retrieval model, a process referred to as the `warm-up' stage.
The aim for training the warm-up model is to be used to initialise the editing model.
This is based on the assumption that these initial clip-caption pairs can still provide decent supervision signal to learn effective representations of clip and caption in the common embedding space, even when derived from timestamp-based initialisation (see Ablation of co-training in Table~\ref{tab:ablation_cotrain}). 
 In the second stage, we adopt a student-teacher framework---co-training two models to edit the clips so as to improve the clip retrieval performance. The teacher model focuses on clip editing, while the student model is trained with the edited clips to assess retrieval performance. This co-training process continues until retrieval performance convergence.

We detail the two stages next.

\myparagraph{Warm up.} The initial clip-caption pairs are used to warm up a clip retrieval model. Our proposed method is model-agnostic, and applicable to any retrieval model with two branches: one for video clips and another for text captions, denoted as $P_V$ and $P_C$ respectively. These branches project the video clip and caption into a common space for similarity measurement. We refer to the embeddings of video clips as $\hat{v}_i = P_V (v_i)$ and the embeddings of captions as $\hat{c}_i = P_C (c_i)$. 

For instance, the retrieval model CLIP4Clip~\cite{clip4clip} is trained with the commonly used symmetric InfoNCE~\cite{infoNCE} loss as follows:
\begin{equation}
   \begin{aligned}
       \mathcal{L} = \frac{1}{\mathcal{B}} \sum_{i\in\mathcal{B}}\log \frac{e^{\sigma(\hat{v}_i,\hat{c}_i)/\tau}}{\sum_{j\in\mathcal{B}}e^{\sigma(\hat{v}_i,\hat{c}_j)/\tau}+e^{{\sigma(\hat{v}_j,\hat{c}_i)}/\tau}},
   \end{aligned}
   \label{eq:initial}
\end{equation}
where $\sigma(\cdot,\cdot)$ calculates the cosine similarity between embeddings $\hat{v}_j$ and $\hat{c}_i$, the $j$-th clip and the $i$-th caption respectively, $\mathcal{B}$ represents the batch size and $\tau$ is the temperature.

Using the warm-up model, we select a subset of training caption-clip pairs with similarity above a threshold $\sigma(\hat{v}_i,\hat{c}_i) > \gamma$. This selected subset of pairs is used to monitor our retrieval performance of the edited clips during co-training. We refer to this as the \textit{control set}. The choice of $\gamma$ is ablated in section~\ref{subsec:ablation}. 

\myparagraph{Co-Training.}  
Inspired by~\cite{BYOL, DINO}, we adopt a student-teacher framework for the co-training of our clip editing and retrieval models.
Note that the choice of `student-teacher' here differs from approaches such as~\cite{buciluǎ2006model, hinton2015distilling} where a smaller student model learns from a larger teacher model by knowledge distillation. 
Instead, as in~\cite{BYOL, DINO}, the student and teacher share the same architecture but with different parameters, and the student is updated by gradient descent while the teacher is updated with the weights of teacher with stop-gradient.

Specifically, we utilize a teacher network as the clip editing model, while the student network is trained for clip retrieval with the edited clips. As a result, the student network learns to improve retrieval performance from the edited clips produced by the teacher network.
Note that both the student and teacher networks share identical architectures and are initialized with the weights from the same warm-up model.

At the start of every epoch, the teacher model is used to edit the clips in the training set to produce more semantically similar clips to the corresponding text captions. 
We divide each clip into equal-length segments and calculate the similarity of each segment to the text caption independently.
We then consider the top $K$ visual segments as potential boundaries of the edited clip. This is based on the assumption that these segments offer the best signal to match the visual to the text caption,
The Top $K$ segments of clip $c_i$ are computed as follows:

\begin{equation}
\label{eq:topk}
E_{TopK}(v_i) = \mathop{\arg \max}_{K} \forall_{v_i^s \in v_i} {\sigma(v_i^s,c_i)},
\end{equation}
where $v_i^s$ refers to a segment within clip $v_i$ and $\sigma(v_i^s,c_i)$ is the cosine similarity between the segment and the caption. Note that $E_{TopK}(v_i)$ is not necessarily consecutive in time, in fact, the $K$ segments are often discrete, capturing the most salient parts corresponding to the caption. 

We next edit the clip by selecting a continuous sub-segment of the clip with a higher similarity to the caption but without diverging too far from the initial clip.
We consider all pairs of segments in the Top $K$ as potential boundaries for editing.
We then find the boundary that maximises the temporal overlap with all other candidate boundaries.Namely,
\begin{equation}
E_{cand}^i =\{[a, b] ; \{a,b\} \subset E_{TopK}(v_i), a < b\}
\end{equation}
\begin{equation}
\begin{split}
E_{v^i} &= \arg \max_{E_{cand_j}^i} \sum_{k}{IoU(E_{cand_j}^i, E_{cand_k}^i)}.
\end{split}
\end{equation}
An example of the editing process can be seen in Figure~\ref{fig:method} right.  In this way, we encourage the edited clip to maximise mutual consensus among the potential candidate boundaries, thus the editing clips are made to be semantically similar to the caption and representative within the initial clip boundary. 


The edited clip-caption pairs are subsequently used to train the student model to improve the retrieval performance.
The student network is trained for one epoch using these edited clips, and the weights of the model are updated by gradient descent from the retrieval model's loss, which is the same as used during the warm-up stage. 
The two models thus diverge as the training progresses.
After each epoch, the student model is assessed against the control set, to determine if the retrieval performance on the control set has improved.
When the performance improves, the student model is deemed to be progressing in the right direction, and the weights of the teacher model are updated via copying to match the weights of the student model.
If the performance of the student model drops, it is trained for another epoch, and assessed again.

This co-training process continues until the retrieval performance on the student network, when evaluated on the control pairs, cannot be improved for $M$ epochs.

\section{Experiments}
\label{sec:exp}

\subsection{Experimental Settings}

\myparagraph{Datasets.} We report results on three datasets.

\noindent \textbf{YouCook2}~\cite{ZhXuCoAAAI18} is an untrimmed video dataset focusing on cooking scenarios sourced from YouTube. Due to unavailable videos, we download 1,186 and 415 videos in train and test sets respectively. Each video contains the annotations of a temporal boundary for cooking steps, along with a corresponding caption. The total numbers of clips in the train and test sets are 9,180 and 3,192. On average, the duration of the video is 5.3 minutes and each clip spans 19.6 seconds.

\noindent \textbf{DiDeMo}~\cite{DiDeMo} is comprised of 10,500 Flickr videos and each video covers a duration of around 30 seconds. A video contains several clips with an average duration of 6.5 seconds and corresponding captions. The start and end times for each clip are annotated by selecting a 5-second segment, though the start and end times are averaged based on multiple annotators, the clip boundary is still noisy for the caption. While DiDeMo was originally designed for the purpose of single video moment retrieval, it has been widely used in video-paragraph retrieval and video corpus moment retrieval. In this paper, the DiDeMo dataset is trimmed into clip-caption pairs and used for the task of clip-retrieval. Following~\cite{clip4clip}, the numbers of videos in the train and test sets are 8,392 and 1,004 respectively. Due to a large degree of temporal overlap in the annotations, we exclude captions with  a temporal overlap higher than 0.6 in the test set for evaluation. As a result, the numbers of clips in train and test sets are 32,724 and 2,578. 

\noindent \textbf{ActivityNet-Captions}~\cite{krishna2017dense} was initially established for the purpose of facilitating dense video captioning. This dataset comprises of 15K videos sourced from YouTube, with an average video length of 117.6 seconds. Both clip boundaries and their corresponding captions are manually annotated. The number of available videos in the training and test sets are 7,502 and 3,676, which results in 27,767 and 13,017 clips respectively. Following~\cite{clip4clip, coot}, we use the `val1' split as a test set for evaluation.

\myparagraph{Evaluation Metrics.} In all results, we report caption-to-clip retrieval results, as these particularly assess our model's ability to retrieve clips from a gallery. The retrieval performance is evaluated by median rank (MedR) $\downarrow$ and recall at top K (R@K) $\uparrow$ for which we report R@1, R@5 and R@10. MedR denotes the median rank position of the corresponding clip when considering all the queries in the test set for retrieval, while R@K is the percentage of true positives being ranked amongst the top K.

\begin{table*}
   \centering
   \resizebox{\linewidth}{!}{
   \begin{tabular}{l|c|c|c|c|c|c|c|c|c|c|c|c|c|c}
   \hline
   \multirow{2}{*}{Retrievel models} & \multirow{2}{*}{Methods} & \multirow{2}{*}{Supervision} &\multicolumn{4}{c|}{YouCook2} &\multicolumn{4}{c|}{DiDeMo} &\multicolumn{4}{c}{ActivityNet-Captions}\\
   \cline{4-15}
    & & & R@1 & R@5 & R@10 & MedR & R@1 & R@5 & R@10 & MedR & R@1 & R@5 & R@10 & MedR \\
   \hline
   \rowcolor{pink!40}
   \cellcolor{white!0} & Baseline  & timestamp & 13.5 & 34.0 & 46.2 & 13 &3.6 &13.8 & 21.5 & 49 & 7.6 & 22.7 & 32.2 & 28 \\
   \cline{2-15}
   \rowcolor{blue!20}
   \cellcolor{white!0}{COOT~\cite{coot}} & Ours & timestamp & \textbf{15.1} & \textbf{36.3} & \textbf{48.1} & \textbf{12} & 3.6 & \textbf{14.7} & \textbf{22.3} & \textbf{47} & \textbf{8.0} & \textbf{22.8} & \textbf{32.8} & \textbf{28} \\
    \cline{2-15}
   \rowcolor{black!20}
   \cellcolor{white!0} & Up-bound & GT & 16 & 38.4 & 51.2 & 10 & 5.8 & 17.98 & 25.83 & 46 & 8.5 & 24.7 & 34.9 & 24 \\
   \hline
   \hline
   \rowcolor{pink!40}
   \cellcolor{white!0} & baseline & timestamp & 28.1	& 57.6 & 70.6 & 4 & 5.2 & 17.0 & 25.3 & 51 & 4.4 & 11.4 & 15.9 & 357 \\
  \cline{2-15}
   \rowcolor{blue!20}
   \cellcolor{white!0}{VideoCLIP~\cite{xu2021videoclip}} & Ours & timestamp & \textbf{28.8} & \textbf{58.4}  & \textbf{71.1} & \textbf{4} & \textbf{5.7} & \textbf{17.6} & \textbf{25.6} & \textbf{49} & \textbf{4.6} & \textbf{11.9} & \textbf{16.5} & \textbf{335} \\
   \cline{2-15}
   \rowcolor{black!20}
   \cellcolor{white!0} & Up-bound & GT & 30.0 & 59.2 & 72.1  & 4 & 5.8 & 18.0 & 26.2 & 45 & 5.2 & 12.8 & 17.8 & 319 \\
    \hline
    \hline
   \rowcolor{pink!40}
    \cellcolor{white!0} & Baseline & timestamp & 10.2 & 27.6 & 38.2 & 21 & 10.8 & 29.8 & 39.8 & 20 & 10.4 & 26.4 & 36.5 & 22\\
  \cline{2-15}
   \rowcolor{blue!20}
    \cellcolor{white!0}{CLIP4Clip~\cite{clip4clip}} & Ours & timestamp & \textbf{10.5} & 27.6 & 38.2 & 21 & 10.8 & 29.6 & \textbf{\textbf{39.9}} & 20 & \textbf{10.7} & 26.4 & 36.5 & 22\\
   \cline{2-15}
   \rowcolor{black!20}
    \cellcolor{white!0} & GT & Up-bound & 10.9 & 28.4 & 39.7 & 18 & 10.8 & 31.2 & 40.4 & 19 & 10.8 & 27.3 & 37.3 & 22 \\
   \hline
   \end{tabular}}
   \caption{Caption-to-clip retrieval performance comparison. The pink rows represent the baseline models trained using single timestamp as supervision. The blue rows refer to the baseline models equipped with our proposed video editing method using timestamps as supervision. The grey rows indicate the upbound retrieval performance of the baseline models which are trained using human-labeled ground truth.}
    \label{tab:retrieval_comparison}
\end{table*}

\myparagraph{Retrieval Models.} Our video clip editing method can be applied to any retrieval model. We employ three widely used state-of-the-art retrieval models to showcase the effectiveness of our proposed method, including COOT~\cite{coot}, VideoCLIP~\cite{xu2021videoclip} and CLIP4Clip~\cite{clip4clip}. 

\begin{itemize}

    \item \textbf{COOT}~\cite{coot} is a hierarchical transformer model for video-text retrieval. The clip-level and sentence-level features are obtained by temporal transformer with attention-based feature aggregation. A contextual transformer is employed to model the interaction between local (clip and sentence) and global (video and paragraph) information for learning video-level and paragraph-level features. Furthermore, a cross-modal cycle-consistency loss is proposed to align the clip and sentence in the joint embedding space. COOT manages to achieve competitive results in YouCook2 and AvtivityNet-Captions datasets. 
    \item \textbf{VideoCLIP}~\cite{xu2021videoclip} is a pre-trained vision-language model using contrastive learning. The key idea of VideoCLIP is to form high-quality positive and negative samples during pre-training. Instead of only using start and end times of clips, the positive text-clip pairs are constructed by shifting the start and end times with temporal overlap varying their length. The negative samples are formed by augmented videos which are more similar to each training batch representing hard negatives. VideoCLIP is pre-trained using the HowTo100M dataset and demonstrates superiority in various tasks, such as video-text retrieval in YouCook2. Different from COOT and VideoClip, which take video features as input.
    \item \textbf{CLIP4Clip}~\cite{clip4clip} is built upon CLIP~\cite{CLIP} and trained in an end-to-end manner with raw videos as input. Both video and text encoders are derived from CLIP, the video embedding is aggregated from frame-level features. CLIP4Clip is one of the state-of-the-art methods in video-text retrieval on the DiDeMo and ActivityNet-Captions datasets, though the reported results are evaluated by video-paragraph retrieval. 
\end{itemize}

\myparagraph{Implementation Details.} 
We use the same hyperparameters and settings as in the published codebases of the three retrieval models, COOT~\cite{coot}, VideoCLIP~\cite{xu2021videoclip} and CLIP4Clip~\cite{clip4clip}. For VideoCLIP and COOT, we follow VideoCLIP to extract visual features by using an S3D model~\cite{miech2020end} pre-trained on HowTo100M~\cite{howto100m} for all the three datasets, which produces a 512 dimensional output. The number of input frames is 30, \emph{i.e.}, one segment contains 30 frames, also equivalent to 1 second. The best-performing student model during training is selected to report retrieval performance and the final teacher model is used for video clip editing. In contrast, CLIP4Clip takes raw videos as input. We sample 1 frame per second with a uniform sampling strategy as the original paper. Mean pooling is employed to aggregate the frame-level features into clip embeddings given its robustness and good performance in different datasets.
The single timestamps for all the three datasets are randomly chosen within the human annotated start and end times. Due to the average clip length in ActivityNet-captions is fairly long, i.e., 37 seconds, the clip editing is prone to conduct dramatic editing. To avoid this, we set an extra IoU threshold between initial and edited clips during the training in ActivityNet. We detailed the ablation of IoU threshold in Section~\ref{subsec:ablation}. 
Note that we edit all the clips in YouCook2 and DiDeMo datasets. 

\begin{figure*}[t]
    \begin{minipage}[t]{0.45\textwidth}
   \includegraphics[width=\textwidth]{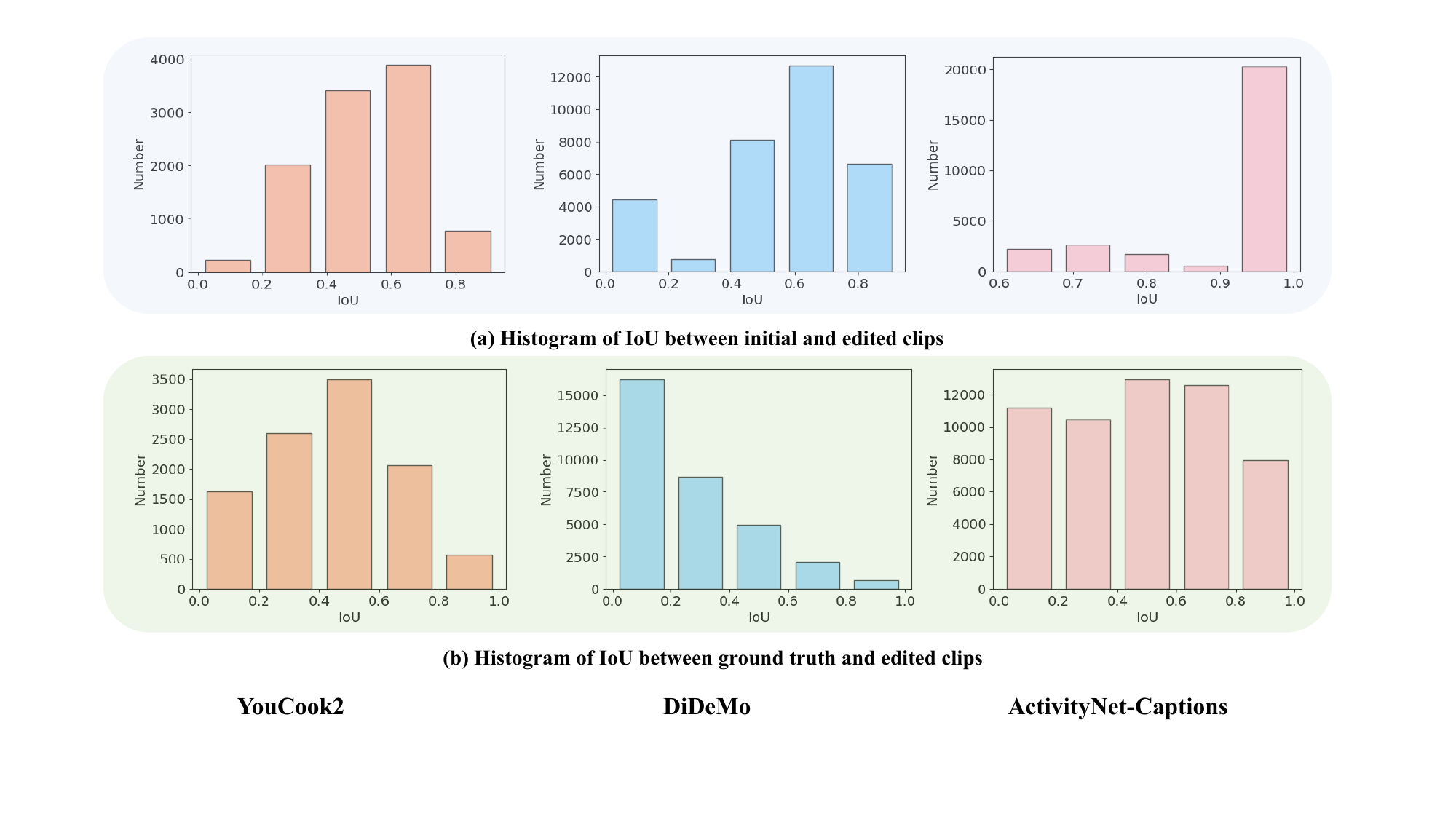}
      \caption{The percentage of clip editing in the training set across three datasets using COOT~\cite{coot}. We present histograms showing the IoU between initial and edited clips (a) and between ground truth and edited clips (b).}
   \label{fig:edit_stats}
   
   \end{minipage}
    \hspace{0.03\textwidth}
    \begin{minipage}[t]{0.48\textwidth}
    \centering
    \includegraphics[width=\textwidth]{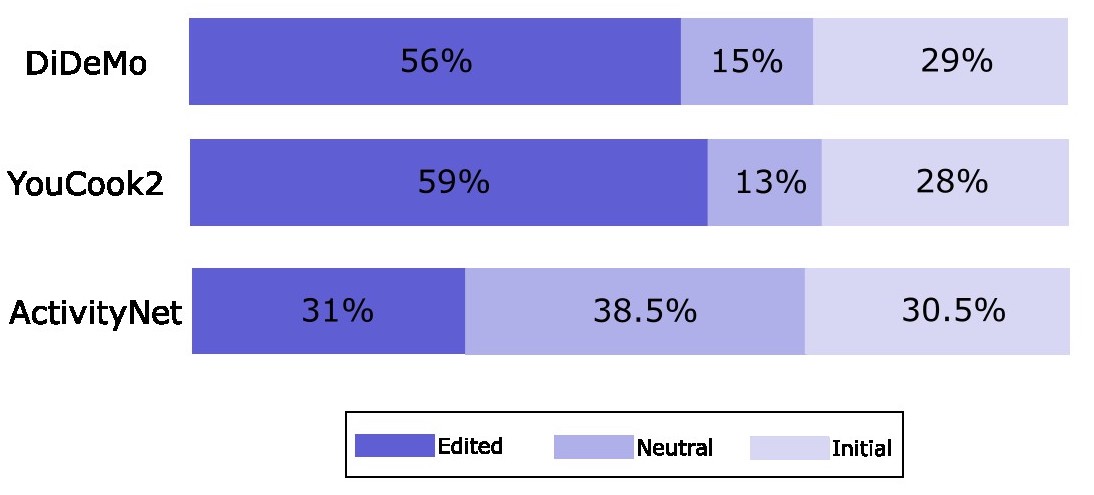}
    \caption{Results from a human study evaluating the correspondence between initial and edited clips to their captions across three video retrieval datasets. The percentage of choices favoring the better clips is reported.}
   \label{fig:human}
   \label{fig:interface}
   \end{minipage}
\end{figure*}

\begin{figure*}[t]
   \includegraphics[width=\textwidth]{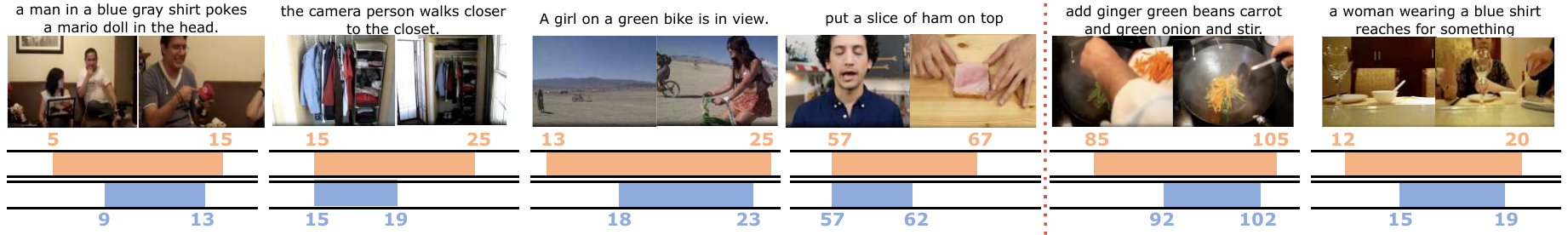}
      \caption{Qualitative examples showing caption-to-clip retrieval with initial ({\color{orange}orange}) and edited ({\color{blue}blue}) clips from different videos, where the timestamps are in seconds. The first four examples show edited clips are superior to initial clips from human perception, and the last two demonstrate the edited clips are inferior to initial clips. Video clips are presented on the project webpage.}
   \label{fig:example}
\end{figure*}

\begin{figure*}[t]
   \includegraphics[width=\textwidth]{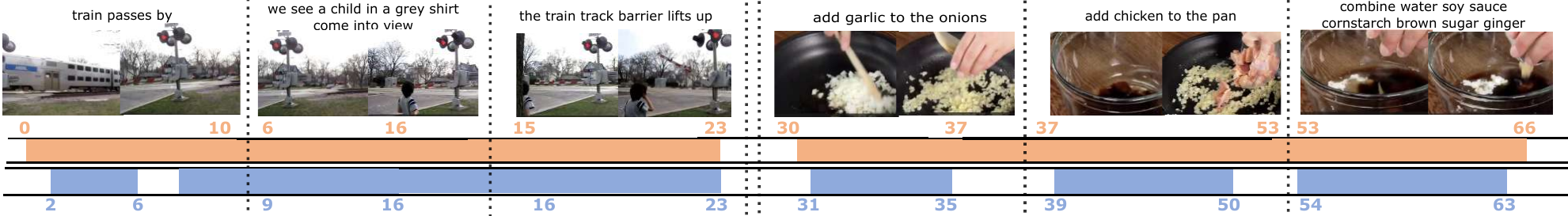}
      \caption{Qualitative examples within two untrimmed videos in DiDeMo and YouCook2 respectively, where the timestamps are in seconds and each with three clips. Initial and edited clips are in {\color{orange}orange} and {\color{blue}blue} respectively. Please refer to video clips presented on the project webpage.}
   \label{fig:longterm_example}
\end{figure*}

\begin{figure*}[t]
   \includegraphics[width=\textwidth]{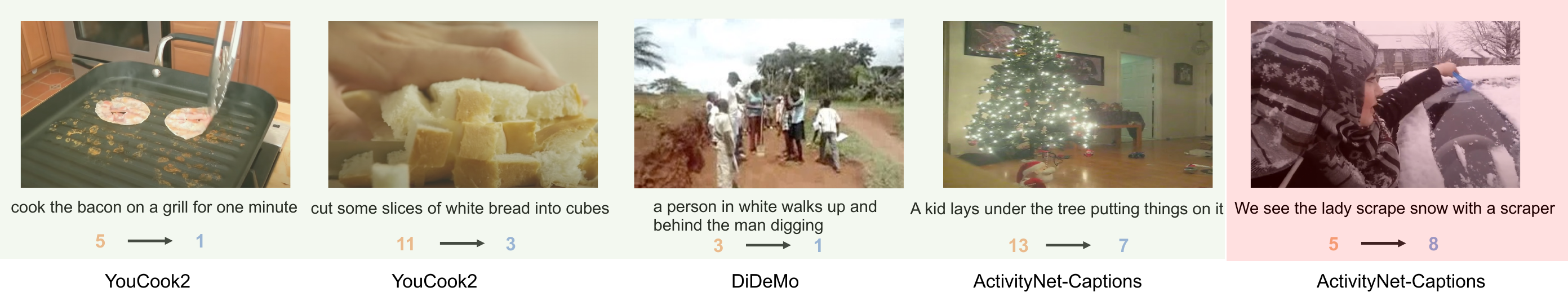}
      \caption{Qualitative examples showcase that our proposed method manages to boost the ranks for clip retrieval. The rank of baseline retrieval model is shown in {\color{orange}orange} and our method is shown in {\color{blue}blue}. The retrieval model is based on COOT.}
   \label{fig:rank_example}
\end{figure*}

\subsection{Result Analysis}
\myparagraph{Clip retrieval performance comparison.} 
As shown in Table~\ref{tab:retrieval_comparison}, we first present caption-to-clip retrieval results of the three retrieval models across all three datasets. For each retrieval model, we have three groups of results presented in different colours. The results of baseline retrieval models using the initial clips $v_i^0$ are shown in pink. Subsequently, we employ the baseline retrieval model with our proposed video clip editing method---results are shown in blue. Note that both of these results are based on timestamps as supervision for training. In addition, we also present the results of each retrieval model trained with ground truth start and end times as upper bound retrieval performance for our method, presented in grey.

After clip editing, it can be observed that our editing module manages to boost the clip retrieval performance consistently across all three retrieval models on all three datasets. For example, in YouCook2, COOT~\cite{coot}+Edit outperforms COOT with a noticeable gain in performance. The MedR is boosted from 13 to 12 with 1 rank improvement, and R@1, R@5 and R@10 are improved by 1.6\%, 2.3\% and 1.9\% respectively. Performance of VideoCLIP is much better than COOT due to large-scale pre-taining on HowTo100M. In contrast, though CLIP4Clip is trained end-to-end, the performance is still inferior than both COOT and VideoCLIP which takes S3D~\cite{miech2020end} features as input. Nevertheless, the S3D feature extractor is a video-based network pre-trained on large-scale HowTo100M dataset, while CLIP4Clip employs CLIP, which is a pre-trained image-based backbone. Finally, we note that compared with retrieval models trained with human labeled ground truth, our proposed method still has a performance gap. It is not surprising as our method only uses weakly supervised single timestamps instead of fully-supervised start and end times for training. 

\myparagraph{Clip editing analysis.} To investigate the effect of clip editing, we analyse the difference 
between initial and edited clips in the training set across the three datasets. As show in Figure~\ref{fig:edit_stats} (a), we plot the histograms of Intersection over Union (IoU) between initial and edited clips. It can be observed that the clip editing in YouCook2 is fairly balanced distributed with different IoUs. Most clips are edited to have IoU of [0.4, 0.8] with the initial clips while only a small portion clips go through significant changes to IoU [0, 0.1] or minor changes to IoU [0.8, 1.0]. In contrast, the clip editing in DiDemo tends to more aggressive than YouCook2. We believe it results from the start and end times annotation in DiDemo based on 5-second segments selection rather than precise annotation as in YouCook2.
The clip editing in ActivityNet-Captions is more conservative by introducing IoU threshold to prevent aggressive editing.
Comparing the edited clips with ground truth, the results in YouCook2 and ActivityNet-captions are balanced across different IoUs with precise start and end times annotation. In contrast, the IoU with DiDeMo tends to be small with noisy annotations.

\myparagraph{Human evaluation.}
A human evaluation was carried out to judge how effective the clip editing is from a human perspective. 
For each dataset, a set of captions from the test set are randomly formed. Each caption corresponds to one initial clip and one edited clip produced by the editing model. Human participants are asked to decide which clip better represents the start and end times of each caption. They were also given a `neutral' option if they felt that both clips matched equally well. Note that the initial and edited clips are randomly shuffled as options A or B, and participants are allowed to watch the clips many times. The result of the DiDeMo dataset evaluation utilizes the CLIP4Clip model, while the analysis of YouCook2 and ActivityNet is conducted using the COOT model. 

As the human perception of minor changes in the clips is not obvious, to ensure that the edited clips were clearly distinguishable from the initial clips, we set a maximum IoU overlap of 0.5 between the initial and edited clips. For DiDeMo, YouCook2 and ActivityNet-Captions, 160 examples were annotated on average 3 times. The results are shown in Figure~\ref{fig:human}. This demonstrates that overall the edited clips are typically judged to be better aligned with the captions than the initial clips, particularly in the case of YouCook2 and DiDeMo.

\myparagraph{Qualitative results.} We present qualitative results of video clip editing from timestamp supervision.
Figure~\ref{fig:example} shows the qualitative examples of our method with initial and edited samples of clips from diverse videos. After editing, the edited clip manages to remove the noisy segments which are irrelevant to the caption. For example, in the first example, the edited clip is shrunken from left and right for 4 and 2 seconds respectively, capturing the key action `poke a Mario doll in the head' more accurately. Similarly, the fourth example is able to curate the narration parts which is not visually aligned with the action `put a slice of ham on top'. We show two failure cases where the edited clips miss part of the information described in the caption or only remove part of the noisy segments. In the fifth example, due to the caption contains many objects, the edited clip missed the `add ginger' part. In the last example, though the noisy segments in the previous part are removed, the latter noisy segments still remain.Figure~\ref{fig:longterm_example} further presents examples of clip editing from two untrimmed videos, where each video contains three consecutive clips. 

In addition, Figure~\ref{fig:rank_example} shows qualitative examples that our method generally improves the ranks compared to the baseline retrieval model. The results further demonstrate that our proposed method manages to improve the correspondence between caption and clip, resulting in improving the video retrieval performance.
Please refer to the videos on the project webpage for more details.

\begin{figure*}[t]
   \centering
   \includegraphics[width=\textwidth]{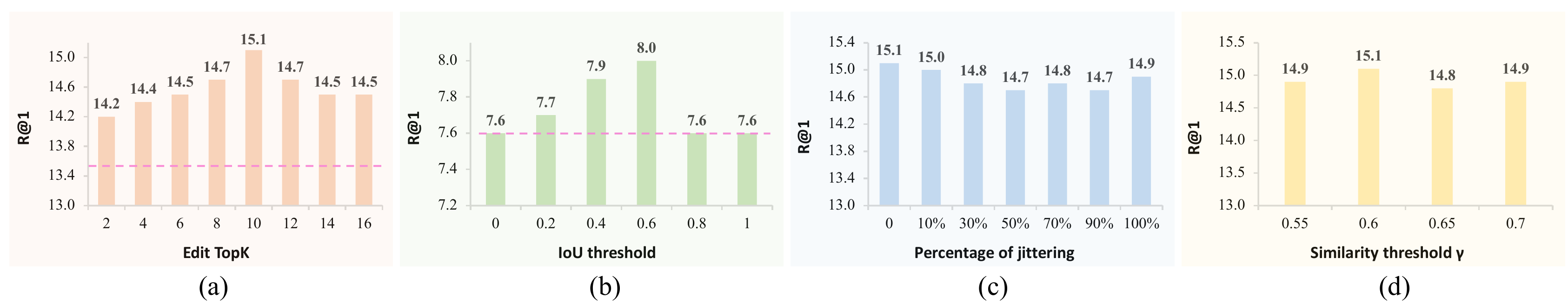}
      \caption{Ablation study. (a) TopK during video clip editing in YouCook2 dataset, (b) IoU threshold during video clip editing in ActivityNet-Captions dataset. (c) Robustness of the proposed method by adding different percentage of jittering to initial clips, (d) Similarity threshold $\gamma$ to form `control set' during video clip editing. All the results are reported based on the retrieval model COOT. The dash line in (a) and (b) refer to the performance of baseline model.}
   \label{fig:ablation_combine}
\end{figure*}

\subsection{Ablation Study}
\label{subsec:ablation}
We conduct ablation studies on how to form the initial
clips, the ending TopK, and the effect co-training, the similarity threshold to select the control set as well as the IoU threshold for our model. Except for the experiments of IoU threshold are conducted on ActivityNet-Captions, all ablations are performed on YouCook2 datasets using COOT.

\myparagraph{Initial clip Formation.} Table~\ref{tab:ablationInit} shows the retrieval results using different strategies to form the initial clips. We consider two groups of strategies: variable length and fixed length. Variable length relies on the relative positions of three consecutive timestamps $t_{i-1}, t_i, t_{i+1}$. 
Fixed length utilises the corresponding timestamp $t_i$ and considers a fixed length segment centred around $t_i$. 
Comparing the two strategies, the best performance is achieved with variable length [$\frac{t_{i-1}+t_{i}}{2},\frac{t_{i}+t_{i+1}}{2}$], which we use for the clip editing experiments.

\myparagraph{Effect of co-training.} As shown in Table~\ref{tab:ablation_cotrain}, we investigate the impact of co-training the student-teacher networks. If the teacher is kept frozen with the weights of warm-up model (3rd row), the retrieval performance drops slightly compared with updating the teacher (2nd row). It can be seen that updating the weights of teacher is beneficial to improve the editing quality and retrieval performance. If the teacher is randomized (row 4th), as expected, the performance becomes worse than the COOT baseline, as the random teacher cannot provide effective clip editing. Furthermore, if we use the student network itself as the teacher (row 5), the performance is improved over the baseline but is inferior to co-training with the updating teacher (row 2). We find that co-training of student and teacher networks is more effective than training and editing using the same network.

\begin{table*}[t]
    \begin{minipage}[t]{0.45\textwidth}
   \resizebox{\linewidth}{!}{
   \begin{tabular}{l|l|c|c|c|c}
   \hline
   \multicolumn{2}{c|}{Initial clip} & R@1 & R@5 & R@10 & MedR \\
   \hline
   \multirow{4}{*}{Variable} & [$t_{i}, t_{i+1}$] & 9.2 & 26.3 & 38.1 & 20 \\
   \cline{2-6}
    & [$t_{i-1}, t_{i}$] & 7.7 & 24.0 & 35.1 & 24 \\
   \cline{2-6}
   & [$t_{i-1}, t_{i+1}$] & 9.0 & 26.6 & 37.7 & 20 \\
   \cline{2-6}
   & [$\frac{t_{i-1}+t_{i}}{2},\frac{t_{i}+t_{i+1}}{2}$] & \textbf{13.5} & \textbf{34.0} & \textbf{46.2} & \textbf{13} \\
   \hline
   \hline
   \multirow{1}{*}{Fixed} & [$t_i-10, t_i+10$] & 11.7 & 31.3 & 43.9 & 14 \\
   \hline
   \end{tabular}}
   \caption{Ablation study for strategies to form the initial clips in YouCook2 dataset. `-$a$' and `+$a$' denote shifting $a$ seconds to the left and right.}
   \label{tab:ablationInit}
   \end{minipage}
    \hspace{0.03\linewidth}
    \begin{minipage}[t]{0.5\textwidth}
    \centering
   \resizebox{\linewidth}{!}{
    \begin{tabular}{c|l|c|c|c|c}
   \hline
    Row & Method & R@1 & R@5 & R@10 & MedR \\
   \hline
    1 & COOT & 13.5 & 34.0 & 46.2 & 13 \\
   \hline
    2 & +Edit (update teacher) & \textbf{15.1} & \textbf{36.3} & \textbf{48.1} & \textbf{12} \\
   \hline
   3 & +Edit (frozen teacher) & 14.7 & 35.6 & 47.6 & 12 \\
   \hline
   4 & +Edit (random teacher) & 13.2 & 33.8 & 46.0 & 13 \\
   \hline
   5 & +Edit (self teacher) & 14.5 & 35.9 & 47.8 & 12 \\
   \hline
   \end{tabular}}
   \caption{Ablation study for co-training. Row 1 presents the baseline COOT model without clip editing. Rows 2 through 5 describe different configurations of the teacher model.}
   \label{tab:ablation_cotrain}
   \end{minipage}
   \vspace*{-12pt}
\end{table*}

\myparagraph{Edit TopK.}
Figure~\ref{fig:ablation_combine}(a), shows the ablation study to investigate impact of TopK (Equation~\ref{eq:topk}) in video clip editing. The best performance is achieved when TopK is assigned to 10. If the $K$ is too small, edited clips tend to only capture the most similar segments while lacking completeness and diversity. If $K$ is too large, it could introduce noisy segments that are not relevant to the caption. Additionally, we notice that our proposed method constantly outperforms the baseline retrieval model COOT regardless of choice of $K$.

\myparagraph{IoU threshold.} Figure~\ref{fig:ablation_combine} (b) shows the ablation study with different choices of IoU threshold to control the extent of video clip editing in the ActivityNet-Captions dataset. By increasing the IoU threshold, video clip editing becomes more and more conservative as clips with aggressive editing below the threshold are abandoned. In ActivityNet-Captions, if we edit all the clips, i.e., IoU threshold equals 0, the retrieval performance is not improved. The R@1 is boosted by increasing the threshold from 0 to 0.6 and the best value is obtained with IoU=0.6, we find that thresholds above 0.6 do not help performance. This is because choosing a large IoU only allows a small portion of clips to be edited, thus reducing the effectiveness of clip editing.

\myparagraph{Robustness.} To show the robustness of the model, we randomly add 1-2 seconds of jittering to the start or end times for the initial clips. As shown in Figure~\ref{fig:ablation_combine} (c), with increasing the percentage of jittering from 0 to 100\%, the retrieval performance slightly changes in terms of R@1, which shows that our proposed method is robust to minor jittering in general. 

\myparagraph{Similarity threshold $\gamma$ to select `control set'.} The control set selected from the training set is used to determine when to update the weights of teacher from student network. As shown in Figure~\ref{fig:ablation_combine} (d), results show that the retrieval performance isn't sensitive to the choice of $\gamma$.

\section{Conclusion}
\label{sec:conclusion}
We have presented a new method to improve clip retrieval performance from single timestamps in untrimmed videos. By leveraging the initial clips from single timestamps, our method edits the initial clips to refine the temporal boundary to improve the retrieval performance. 
For future work, we wish to explore clip editing for other video understanding tasks, avoiding typical biases in manually trimmed videos, as previously shown in~\cite{Alwassel_2018_ECCV,Moltisanti_2017_ICCV}.

\noindent \textbf{Acknowledgement.} This paper used publicly available datasets. Project supported by EPSRC Program Grant Visual AI (EP/T028572/1) and EPSRC UMPIRE (EP/T004991/1). 
K. Flanagan is supported by UKRI (CDT in Interactive AI Grant ref EP/S022937/1) \& Qinetiq Ltd via studentship CON11954. We also asknowledge the use of the HPC BC4 Cluster of the University of Bristol.

%
%
\bibliographystyle{splncs04}
\bibliography{main}

\begin{thebibliography}{10}
\providecommand{\url}[1]{\texttt{#1}}
\providecommand{\urlprefix}{URL }
\providecommand{\doi}[1]{https://doi.org/#1}

\bibitem{Alwassel_2018_ECCV}
Alwassel, H., Heilbron, F.C., Escorcia, V., Ghanem, B.: Diagnosing error in temporal action detectors. In: Proceedings of the European Conference on Computer Vision. pp. 256--272 (2018)

\bibitem{DiDeMo}
Anne~Hendricks, L., Wang, O., Shechtman, E., Sivic, J., Darrell, T., Russell, B.: Localizing moments in video with natural language. In: Proceedings of the IEEE international conference on computer vision. pp. 5803--5812 (2017)

\bibitem{anne2017localizing}
Anne~Hendricks, L., Wang, O., Shechtman, E., Sivic, J., Darrell, T., Russell, B.: Localizing moments in video with natural language. In: Proceedings of the IEEE International Conference on Computer Vision. pp. 5803--5812 (2017)

\bibitem{ashutosh2023you}
Ashutosh, K., Girdhar, R., Torresani, L., Grauman, K.: What you say is what you show: Visual narration detection in instructional videos. arXiv preprint arXiv:2301.02307  (2023)

\bibitem{buciluǎ2006model}
Buciluǎ, C., Caruana, R., Niculescu-Mizil, A.: Model compression. In: Proceedings of the ACM SIGKDD International Conference on Knowledge Discovery and Data Mining. pp. 535--541 (2006)

\bibitem{DINO}
Caron, M., Touvron, H., Misra, I., J{\'e}gou, H., Mairal, J., Bojanowski, P., Joulin, A.: Emerging properties in self-supervised vision transformers. In: Proceedings of the IEEE/CVF International Conference on Computer Vision. pp. 9650--9660 (2021)

\bibitem{cheng2023vindlu}
Cheng, F., Wang, X., Lei, J., Crandall, D., Bansal, M., Bertasius, G.: Vindlu: A recipe for effective video-and-language pretraining. In: Proceedings of the IEEE/CVF Conference on Computer Vision and Pattern Recognition. pp. 10739--10750 (2023)

\bibitem{cui2022video}
Cui, R., Qian, T., Peng, P., Daskalaki, E., Chen, J., Guo, X., Sun, H., Jiang, Y.G.: Video moment retrieval from text queries via single frame annotation. In: Proceedings of the International ACM SIGIR Conference on Research and Development in Information Retrieval. pp. 1033--1043 (2022)

\bibitem{flanagan2023learning}
Flanagan, K., Damen, D., Wray, M.: Learning temporal sentence grounding from narrated egovideos. arXiv preprint arXiv:2310.17395  (2023)

\bibitem{fragomeni2022contra}
Fragomeni, A., Wray, M., Damen, D.: Contra:({C}on)text ({T}ra)nsformer for cross-modal video retrieval. In: Proceedings of the Asian Conference on Computer Vision. pp. 3481--3499 (2022)

\bibitem{gabeur2020multi}
Gabeur, V., Sun, C., Alahari, K., Schmid, C.: Multi-modal transformer for video retrieval. In: Proceedings of the European Conference on Computer Vision. pp. 214--229 (2020)

\bibitem{gao2017tall}
Gao, J., Sun, C., Yang, Z., Nevatia, R.: Tall: Temporal activity localization via language query. In: Proceedings of the IEEE International Conference on Computer Vision. pp. 5267--5275 (2017)

\bibitem{coot}
Ging, S., Zolfaghari, M., Pirsiavash, H., Brox, T.: Coot: Cooperative hierarchical transformer for video-text representation learning. In: Advances in Neural Information Processing Systems. pp. 22605--22618 (2020)

\bibitem{XPool}
Gorti, S.K., Vouitsis, N., Ma, J., Golestan, K., Volkovs, M., Garg, A., Yu, G.: X-pool: Cross-modal language-video attention for text-video retrieval. In: Proceedings of the IEEE/CVF Conference on Computer Vision and Pattern Recognition. pp. 5006--5015 (2022)

\bibitem{BYOL}
Grill, J.B., Strub, F., Altch{\'e}, F., Tallec, C., Richemond, P., Buchatskaya, E., Doersch, C., Avila~Pires, B., Guo, Z., Gheshlaghi~Azar, M., et~al.: Bootstrap your own latent-a new approach to self-supervised learning. Advances in Neural Information Processing Systems pp. 21271--21284 (2020)

\bibitem{han2022temporal}
Han, T., Xie, W., Zisserman, A.: Temporal alignment networks for long-term video. In: Proceedings of the IEEE/CVF Conference on Computer Vision and Pattern Recognition. pp. 2906--2916 (2022)

\bibitem{hinton2015distilling}
Hinton, G., Vinyals, O., Dean, J.: Distilling the knowledge in a neural network. arXiv preprint arXiv:1503.02531  (2015)

\bibitem{hu2022lightweight}
Hu, F., Chen, A., Wang, Z., Zhou, F., Dong, J., Li, X.: Lightweight attentional feature fusion: A new baseline for text-to-video retrieval. In: Proceedings of the European Conference on Computer Vision. pp. 444--461 (2022)

\bibitem{cross-modal-noisy}
Huang, Z., Niu, G., Liu, X., Ding, W., Xiao, X., Wu, H., Peng, X.: Learning with noisy correspondence for cross-modal matching. In: Advances in Neural Information Processing Systems. pp. 29406--29419 (2021)

\bibitem{jiang2018mentornet}
Jiang, L., Zhou, Z., Leung, T., Li, L.J., Fei-Fei, L.: Mentornet: Learning data-driven curriculum for very deep neural networks on corrupted labels. In: International Conference on Machine Learning. pp. 2304--2313 (2018)

\bibitem{krishna2017dense}
Krishna, R., Hata, K., Ren, F., Fei-Fei, L., Carlos~Niebles, J.: Dense-captioning events in videos. In: Proceedings of the International Conference on Computer Vision. pp. 706--715 (2017)

\bibitem{lee2021learning}
Lee, P., Byun, H.: Learning action completeness from points for weakly-supervised temporal action localization. In: Proceedings of the IEEE/CVF International Conference on Computer Vision. pp. 13648--13657 (2021)

\bibitem{lei2021less}
Lei, J., Li, L., Zhou, L., Gan, Z., Berg, T.L., Bansal, M., Liu, J.: Less is more: Clipbert for video-and-language learning via sparse sampling. In: Proceedings of the IEEE/CVF Conference on Computer Vision and Pattern Recognition. pp. 7331--7341 (2021)

\bibitem{kevin2022egovlp}
Lin, K.Q., Wang, A.J., Soldan, M., Wray, M., Yan, R., Xu, E.Z., Gao, D., Tu, R., Zhao, W., Kong, W., et~al.: Egocentric video-language pretraining. In: Advances in Neural Information Processing Systems. pp. 7575--7586 (2022)

\bibitem{liu2019use}
Liu, Y., Albanie, S., Nagrani, A., Zisserman, A.: Use what you have: Video retrieval using representations from collaborative experts. In: Proceedings of the British Machine Vision Conference. pp. 279--295 (2019)

\bibitem{clip4clip}
Luo, H., Ji, L., Zhong, M., Chen, Y., Lei, W., Duan, N., Li, T.: Clip4clip: An empirical study of clip for end to end video clip retrieval and captioning. Neurocomputing  \textbf{508},  293--304 (2022)

\bibitem{ma2020sf}
Ma, F., Zhu, L., Yang, Y., Zha, S., Kundu, G., Feiszli, M., Shou, Z.: Sf-net: Single-frame supervision for temporal action localization. In: Proceedings of the European Conference on Computer Vision. pp. 420--437 (2020)

\bibitem{miech2020end}
Miech, A., Alayrac, J.B., Smaira, L., Laptev, I., Sivic, J., Zisserman, A.: End-to-end learning of visual representations from uncurated instructional videos. In: Proceedings of the IEEE/CVF Conference on Computer Vision and Pattern Recognition. pp. 9879--9889 (2020)

\bibitem{howto100m}
Miech, A., Zhukov, D., Alayrac, J.B., Tapaswi, M., Laptev, I., Sivic, J.: Howto100m: Learning a text-video embedding by watching hundred million narrated video clips. In: Proceedings of the IEEE/CVF International Conference on Computer Vision. pp. 2630--2640 (2019)

\bibitem{mithun2019weakly}
Mithun, N.C., Paul, S., Roy-Chowdhury, A.K.: Weakly supervised video moment retrieval from text queries. In: Proceedings of the IEEE/CVF Conference on Computer Vision and Pattern Recognition. pp. 11592--11601 (2019)

\bibitem{Moltisanti_2017_ICCV}
Moltisanti, D., Wray, M., Mayol-Cuevas, W., Damen, D.: Trespassing the boundaries: Labeling temporal bounds for object interactions in egocentric video. In: Proceedings of the IEEE International Conference on Computer Vision. pp. 2886--2894 (2017)

\bibitem{infoNCE}
Oord, A.v.d., Li, Y., Vinyals, O.: Representation learning with contrastive predictive coding. arXiv preprint arXiv:1807.03748  (2018)

\bibitem{egovlpv2}
Pramanick, S., Song, Y., Nag, S., Lin, K.Q., Shah, H., Shou, M.Z., Chellappa, R., Zhang, P.: Egovlpv2: Egocentric video-language pre-training with fusion in the backbone. In: Proceedings of the IEEE/CVF International Conference on Computer Vision. pp. 5285--5297 (2023)

\bibitem{CLIP}
Radford, A., Kim, J.W., Hallacy, C., Ramesh, A., Goh, G., Agarwal, S., Sastry, G., Askell, A., Mishkin, P., Clark, J., et~al.: Learning transferable visual models from natural language supervision. In: International Conference on Machine Learning. pp. 8748--8763 (2021)

\bibitem{reed2014training}
Reed, S., Lee, H., Anguelov, D., Szegedy, C., Erhan, D., Rabinovich, A.: Training deep neural networks on noisy labels with bootstrapping. arXiv preprint arXiv:1412.6596  (2014)

\bibitem{ren2018learning}
Ren, M., Zeng, W., Yang, B., Urtasun, R.: Learning to reweight examples for robust deep learning. In: International Conference on Machine Learning. pp. 4334--4343 (2018)

\bibitem{shvetsova2022everything}
Shvetsova, N., Chen, B., Rouditchenko, A., Thomas, S., Kingsbury, B., Feris, R.S., Harwath, D., Glass, J., Kuehne, H.: Everything at once-multi-modal fusion transformer for video retrieval. In: Proceedings of the IEEE/CVF Conference on Computer Vision and Pattern Recognition. pp. 20020--20029 (2022)

\bibitem{song2022learning}
Song, H., Kim, M., Park, D., Shin, Y., Lee, J.G.: Learning from noisy labels with deep neural networks: A survey. IEEE Transactions on Neural Networks and Learning Systems pp. 1--19 (2022)

\bibitem{tanaka2018joint}
Tanaka, D., Ikami, D., Yamasaki, T., Aizawa, K.: Joint optimization framework for learning with noisy labels. In: Proceedings of the IEEE Conference on Computer Vision and Pattern Recognition. pp. 5552--5560 (2018)

\bibitem{wray2019fine}
Wray, M., Larlus, D., Csurka, G., Damen, D.: Fine-grained action retrieval through multiple parts-of-speech embeddings. In: Proceedings of the IEEE/CVF International Conference on Computer Vision. pp. 450--459 (2019)

\bibitem{wu2023cap4video}
Wu, W., Luo, H., Fang, B., Wang, J., Ouyang, W.: Cap4video: What can auxiliary captions do for text-video retrieval? In: Proceedings of the IEEE/CVF Conference on Computer Vision and Pattern Recognition. pp. 10704--10713 (2023)

\bibitem{xiao2015learning}
Xiao, T., Xia, T., Yang, Y., Huang, C., Wang, X.: Learning from massive noisy labeled data for image classification. In: Proceedings of the IEEE Conference on Computer Vision and Pattern Recognition. pp. 2691--2699 (2015)

\bibitem{xu2021videoclip}
Xu, H., Ghosh, G., Huang, P.Y., Okhonko, D., Aghajanyan, A., Metze, F., Zettlemoyer, L., Feichtenhofer, C.: Videoclip: Contrastive pre-training for zero-shot video-text understanding. In: Proceedings of the Conference on Empirical Methods in Natural Language Processing. pp. 6787--6800 (2021)

\bibitem{Msr-vtt}
Xu, J., Mei, T., Yao, T., Rui, Y.: Msr-vtt: A large video description dataset for bridging video and language. In: Proceedings of the IEEE Conference on Computer Vision and Pattern Recognition. pp. 5288--5296 (2016)

\bibitem{xue2023clipvip}
Xue, H., Sun, Y., Liu, B., Fu, J., Song, R., Li, H., Luo, J.: {CLIP}-vip: Adapting pre-trained image-text model to video-language alignment. In: International Conference on Learning Representations (2023)

\bibitem{yi2019probabilistic}
Yi, K., Wu, J.: Probabilistic end-to-end noise correction for learning with noisy labels. In: Proceedings of the IEEE/CVF Conference on Computer Vision and Pattern Recognition. pp. 7017--7025 (2019)

\bibitem{zhang2018cross}
Zhang, B., Hu, H., Sha, F.: Cross-modal and hierarchical modeling of video and text. In: Proceedings of the European Conference on Computer Vision. pp. 374--390 (2018)

\bibitem{lavila}
Zhao, Y., Misra, I., Kr{\"a}henb{\"u}hl, P., Girdhar, R.: Learning video representations from large language models. In: Proceedings of the IEEE/CVF Conference on Computer Vision and Pattern Recognition. pp. 6586--6597 (2023)

\bibitem{ZhXuCoAAAI18}
Zhou, L., Xu, C., Corso, J.J.: Towards automatic learning of procedures from web instructional videos. In: AAAI Conference on Artificial Intelligence. pp. 7590--7598 (2018)

\end{thebibliography}
\end{document}